# LOOK, RECITE, THEN ANSWER: ENHANCING VLM PERFORMANCE VIA SELF-GENERATED KNOWLEDGE HINTS

**Xisheng Feng**[1]

[1] School of Plant Protection, Anhui Agricultural University

## ABSTRACT

Vision-Language Models (VLMs) exhibit significant performance plateaus in specialized domains like precision agriculture, primarily due to "Reasoning-Driven Hallucination" where linguistic priors override visual perception. A key bottleneck is the "Modality Gap": visual embeddings fail to reliably activate the fine-grained expert knowledge already encoded in model parameters.

We propose "Look, Recite, Then Answer," a parameter-efficient framework that enhances VLMs via self-generated knowledge hints while keeping backbone models frozen. The framework decouples inference into three stages: (1) Look generates objective visual descriptions and candidate sets; (2) Recite employs a lightweight 1.7B router to transform visual cues into targeted queries that trigger candidate-specific parametric knowledge; (3) Answer performs parallel evidence alignment between descriptions and recited knowledge to select the most consistent label.

On AgroBench, our method achieves state-of-the-art results, improving Weed Identification accuracy by 23.52% over Qwen2-VL-72B and surpassing GPT-4o without external search overhead. This modular design mitigates hallucinations by transforming passive perception into active, controllable knowledge retrieval
 Our code and model will be available at
`https://github.com/cxk66666666666/look-recite-answer-vlm`.

## 1 INTRODUCTION

Vision-Language Models (VLMs) have advanced by synergizing visual representations with linguistic reasoning. However, they exhibit significant performance plateaus in specialized, interdisciplinary domains like precision agriculture. Empirical analyses from AgroBench reveal a critical failure mode: "Reasoning-Driven Hallucination." (Shinoda et al., 2025). When models lack fine-grained domain grounding, strong linguistic priors often overwhelm visual perception, compelling the fabrication of evidence—such as describing non-existent lesion patterns—to rationalize speculative predictions (Shinoda et al., 2025; Lin et al., 2024). This underscores a structural deficiency: while domain knowledge is latent within the model parameters, the coupling of perception and reasoning often leads to confirmation bias rather than factual deduction (Lu et al., 2025).

To address knowledge deficits, Retrieval-Augmented Generation (RAG) typically grounds outputs in external sources (Gao et al., 2023). However, external web retrieval incurs latency and stochastic noise (Zhang et al., 2024). Conversely, NLP research (e.g., RECITE, ZEROSEARCH) demonstrates that leveraging the LLM itself as a parametric knowledge base yields higher consistency (Sun et al., 2023; Sun et al., 2025). Yet, transferring this paradigm to VLMs faces a specific "Modality Gap": unlike discrete text tokens, standard continuous visual embeddings often fail to precisely activate the specific encyclopedic retrieval pathways required for fine-grained classification, (Liang et al., 2022; Chen et al., 2025) rendering internal expert knowledge inefficiently accessible during visual inference.

During our multi-agent system development, a misconfigured agent bypassed external search and instead retrieved content from its LLM backbone, surprisingly yielding superior diagnostic accuracy. This anomaly suggested a powerful hypothesis: in complex visual reasoning scenarios, models do not necessarily require external search engines; rather, they require a structured mechanism to activate the latent knowledge they already possess.

Building on this insight, we propose "LOOK, RECITE, THEN ANSWER," a novel inference framework enhancing VLM performance via Self-Generated Knowledge Hints. Adopting a parameter-efficient design, we maintain the entire VLM backbone in a frozen state while training a lightweight module to bridge the modality gap. The process structurally decouples into: (1) Look—generating objective, structured visual descriptions via the frozen VLM to anchor perception; (2) Recite—employing a trainable "Candidate-Guided Knowledge Router" that contextualizes visual cues with query candidates to trigger discriminative parametric knowledge retrieval; and (3) Answer—aligning recited facts with verified visual evidence via a frozen reasoning module for final verification.

Our framework yields state-of-the-art performance on the AgroBench suite. By unlocking internal knowledge, we achieve significant improvements over the `Qwen2-VL-72B` baseline—specifically increasing accuracy by 23.52% in Weed Identification—and surpass leading proprietary models such as GPT-4o without external search overhead. Our contributions are:

- Structural Decoupling: A framework that mitigates hallucinations by enforcing a strict separation between visual perception and deductive reasoning.

- Bridging the Modality Gap: A Candidate-Guided Router that translates visual evidence into targeted queries, transforming passive perception into active, learnable knowledge retrieval.

- Efficiency and Modularity: A frozen backbone strategy requiring minimal computational overhead, ensuring modular integration with evolving foundation models.

## 2 RELATED WORK

### 2.1 STRUCTURAL HALLUCINATIONS AND KNOWLEDGE GAPS IN DOMAIN VLMS

While Vision-Language Models (VLMs) demonstrate robust capabilities in general contexts, their deployment in specialized domains is hindered by structural cognitive deficits. Shinoda et al. (2025) utilized AgroBench, an expert-annotated benchmark, to empirically dissect these failures. Their analysis identified "Lack of Knowledge" as the primary error source (51.92%), where models correctly perceive visual entities but fail to recall specific domain facts (e.g., taxonomy or treatment protocols). More critically, they uncovered "Perceptual Errors" (32.69%) that frequently manifest as reasoning-driven hallucinations.For example, to rationalize an incorrect disease prediction, a model often fabricates visual evidence—such as describing "purple lesions" where none exist—to align with its internal priors. Furthermore, models are prone to "Shortcut Errors" (7.79%), selecting an answer without rigorously comparing it against other plausible candidates. These findings indicate that standard end-to-end VLMs struggle to disentangle visual perception from reasoning (Chen et al., 2024; Zhou et al., 2025), necessitating a framework that structurally decouples these processes.

### 2.2 THE SUPREMACY OF SIMULATED INTERNAL RETRIEVAL

To address knowledge gaps without the latency and unpredictability of external Retrieval-Augmented Generation (RAG), research has pivoted towards leveraging Large Language Models (LLMs) as internal knowledge bases. Zhengbao Sun et al. (2023) introduced RECITE, demonstrating that LLMs possess "fuzzy memorization" accessible only through an intermediate recitation step, achieving state-of-the-art results on closed-book QA tasks. Recently, Hongyi Sun et al. (2025) advanced this concept with ZEROSEARCH, a framework that replaces real search engines with simulated search via an LLM. They proved that a simulated retrieval module, when trained with a noise-injecting curriculum, not only eliminates API costs but also outperforms real search engines by providing controllable, high-quality evidence. This validates our strategy: utilizing the model's own parametric knowledge via "Self-Generated Hints" is a superior approach for ensuring the coherence and relevance of retrieved information in reasoning tasks.

## 2.3 BRIDGING THE GAP VIA CANDIDATE-GUIDED KNOWLEDGE RETRIEVAL

A critical limitation of both RECITE and ZEROSEARCH is their confinement to the unimodal text domain, relying on textual queries to trigger knowledge retrieval. This creates a "Modality Gap" for VLMs (Liang et al., 2022): raw visual features cannot directly drive the sophisticated internal retrieval mechanisms described in ZEROSEARCH. Our work explicitly bridges this gap. We introduce a specialized "Candidate-Guided Knowledge Retrieval" model that synthesizes two critical inputs: (1) the structured visual description generated by the VLM's "Look" phase, and (2) the specific candidate set extracted from the user's inquiry (e.g., "Apple Scab vs. Powdery Mildew"). This router acts as a navigator, explicitly directing the "Recite" module to retrieve parametric symptom descriptions for each candidate. Finally, a frozen reasoning module performs the critical alignment, rigorously cross-referencing the recited knowledge against the verified visual features. This mechanism forces the system to execute an "Evidence-Based Verification" process rather than relying on "Speculative Inference" (Zhang et al., 2025), structurally preventing the shortcut learning and hallucination loops identified in AgroBench.

## 3 METHODOLOGY: LOOK, RECITE, THEN ANSWER

### 3.1 FRAMEWORK OVERVIEW

The goal is to answer a user query $Q$ about an input image $I$ by selecting the correct answer $c*$ from a candidate set $C = c_1, ..., c_n$. Traditional end-to-end VLMs directly model (Alayrac et al., 2022)

$$c* = \underset{c \in C}{argmax} P(c \mid I, Q)$$

so that the probability of each candidate $c$ depends simultaneously on raw visual features and linguistic context, tightly coupling perception and reasoning.

We instead introduce intermediate representations $D$ and $K$ and approximate

$$P(c \mid I, Q) \approx P(c \mid D, K_c)$$

Where $D = f_{VLM}(I, Q; \theta_{VLM}^{frozen})$ is a structured visual description produced by a frozen VLM, and $K_c$ is the candidate-specific knowledge retrieved for $c$, drawn from a collection $K = K_1, ..., K_n$ with $K_i$ aligned to $c_i$. This reformulation replaces direct conditioning on pixels with conditioning on explicit visual descriptions and candidate-guided knowledge, and is based on the insight that modern LMMs are bottlenecked less by missing knowledge than by the lack of mechanisms to reliably activate the knowledge already encoded in their parameters (Shi et al., 2025)

To operationalize this idea, we design a three-stage pipeline :

- **Look**: A frozen VLM generates a structured description $D$ and candidate set $C$, providing an objective perceptual and hypothesis space.

- **Recite**: For each candidate $c_i \in C$, a lightweight trainable router generates a minimal, neutral query $q_i$ that activates candidate-specific parametric knowledge $K_i$ from a frozen LLM.

- **Answer**: A frozen reasoning module jointly considers $D$ and $K_i$ performs evidence alignment for each candidate, and predicts

$$c = \underset{c_i \in C}{argmax} P(c_i \mid D, K_i).$$

This architecture yields a clear decoupling between perception and reasoning, while remaining parameter-efficient: only the 1.7B-parameter router is updated, and both the VLM and LLM backbones remain frozen (Zhang et al., 2024; Jiang et al., 2024). It transforms passive end-to-end inference into active, controllable knowledge activation.

## 3.2 LOOK: STRUCTURED PERCEPTION AND CANDIDATE SPACE

The Look stage constructs (i) an objective textual description of the image and (ii) an explicit candidate space for subsequent reasoning. A frozen VLM $f_{VLM}$ first produces.

$$D = fVLM(I, Q; \theta_{VLM}^{frozen}),$$

where prompts are designed to elicit observable attributes—such as shapes, colors, textures, and spatial distributions—without requesting any classification or diagnosis (Luo et al., 2024; Zhou et al., 2025). This prompt design, combined with freezing $\theta_{VLM}^{frozen}$, prevents perception from drifting toward hypothesized labels and preserves the objectivity of $D$. (Shi et al., 2025)

The candidate set $C$ is then defined as

$$C = C_{user} \quad (if\ the\ query\ Q\ provides\ explicit options)$$

$$f_{nominate}(D, Q; \theta_{VLM}^{frozen}) \quad (otherwise).$$

In closed-ended QA, user-provided choices (e.g., multiple-choice options) are parsed into $C_{user}$. In open-ended QA, the frozen VLM nominates a small set of plausible candidates from its parametric knowledge, conditioned on $D$ and $Q$ (Li et al., 2023). In this setting, the VLM plays the role of an *observer and nominator*, not a final decision-maker: the nominated set may be imperfect, but will be refined by downstream knowledge retrieval and verification.

By the end of the Look stage, we obtain a pair $(D, C)$ that defines a bounded and semantically meaningful search space for subsequent retrieval.

## 3.3 RECITE: CANDIDATE-GUIDED PARAMETRIC KNOWLEDGE ACTIVATION

The Recite stage addresses the modality gap between continuous visual representations and the discrete text queries needed to activate LLM-internal knowledge. In standard VLM pipelines, linearized descriptions like "the leaf has lesions" are often too generic to trigger entity-specific knowledge pathways (e.g., "Apple Scab" vs. "Apple Rust") within the LLM (Petroni et al., 2021; Mallen et al., 2023).

We introduce a candidate-guided router $f_{router}$ (implemented with Qwen3 − 1.7B − Base) that, for each candidate $c_i$, produces a minimal, neutral retrieval query:

$$q_i = f_{router}([D; c_i; Q]; \theta), \forall c_i \in C$$

The router is deliberately constrained in scope: it does not enumerate fine-grained symptom dimensions or suggest any specific visual attributes; instead, it simply requests concise, entity-specific symptom knowledge in the appropriate context (e.g., "on apple leaves") inferred from $D$. For example, given a leaf-level description $D$ and candidates $C =$ Apple Scab, Apple Rust:

For $c_1 = $ *Apple Scab,* the router might generate
*"Retrieve symptom knowledge of Apple Scab on apple leaves."*
For $c_2 = $ *Apple Rust,* it might generate
*"Retrieve symptom knowledge of Apple Rust on apple leaves."*

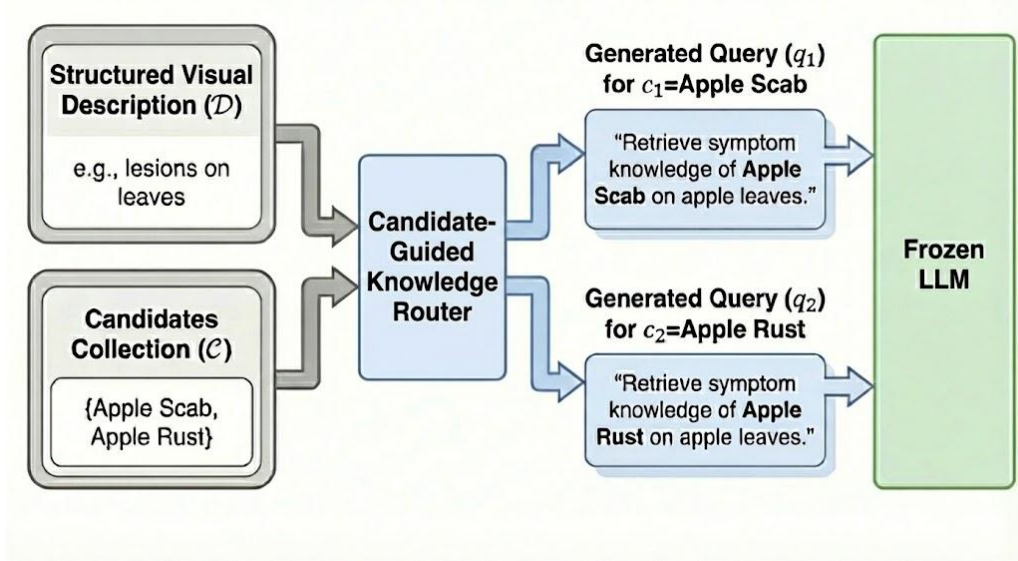

Thus, the router acts as a knowledge indexer: it uses $[D; c_i; Q]$ only to decide *which* entity's knowledge, and in *which organ-level context*, should be retrieved, without embedding any hypothesized features into the query

Each query $q_i$ is then passed to a frozen LLM to retrieve candidate-specific knowledge from its parametric memory $\mathcal{M}_{\text{param}}$:

$$K_i = LLM_{frozen}(q_i; M_{param}).$$

Following the paradigm of Recitation-Augmented Language Models (RECITE), the LLM outputs a short, textbook-style description $K_i$ for entity $c_i$, such as the canonical symptomatology and progression patterns documented in encyclopedic sources.

Collecting all outputs yields a candidate-guided knowledge set

$$K = K_1, \ldots, K_n, K_i = LLM_{frozen}(f_{router}([D; c_i; Q]; \theta)).$$

Here, RECITE (all caps) refers to the external work "Recitation-Augmented Language Models," which uses multiple independent recitations and majority voting to improve answer robustness; our Recite stage instead uses candidate-wise retrieval but is inspired by the same idea of grounding answers in explicit recited knowledge rather than raw parametric priors.

### 3.4 ANSWER: MULTI-CANDIDATE PARALLEL VERIFICATION AND EVIDENCE ALIGNMENT

The Answer stage implements a structured decision process by performing parallel verification across all candidates using a frozen reasoning module. For each $c_i \in C$, we estimate its probability given the visual description and retrieved knowledge as

$$P(c_i \mid I, Q) \approx P(c_i \mid D, K_i) = \prod P(c_{i,t} \mid c_{i,<t}, D, K_i),$$

where $c_{i,t}$ is the $t$-th token of the candidate label.

Rather than re-encoding the image, the reasoning module is prompted to assess the compatibility between $D$ and $K_i$ for each candidate $c_i$. Concretely, it is asked to: (1) identify which attributes in $D$ support or match diagnostic criteria in $K_i$; (2) detect contradictions where $K_i$ mentions key features that are absent or negated in $D$; and (3) judge whether the evidence described in $D$ is sufficiently covered by $K_i$ to support $c_i$. The output is a scalar consistency score, which we interpret as $P(c_i \mid D, K_i)$, optionally accompanied by a short textual rationale.

This design is related to the self-consistency mechanism in RECITE. RECITE generates multiple independent recitations for the same question and aggregates their answers (e.g., via majority vote) to reduce hallucinations. In contrast, our framework generates a single recitation per candidate but evaluates all candidates in parallel: each $(D, K_i)$ pair forms an independent "evidence chain," and the model selects the answer whose chain exhibits the strongest internal consistency. Even if the Look stage proposes an imperfect candidate set, comparing all $P(c_i \mid D, K_i)$ allows the Answer stage to automatically down-weight inconsistent candidates and pick the most evidence-supported one.

The final prediction is

$$\hat{c} = \underset{c_i \in C}{argmax} P(c_i \mid D, K_i).$$

Structurally, this mitigates three major failure modes: (1) reasoning-driven hallucinations, by preventing reasoning from directly overriding perception; (2) shortcut learning, by enforcing explicit evidence alignment between $D$ and $K_i$; and (3) confirmation bias, by requiring parallel evaluation of all candidates rather than testing a single favored hypothesis.

### 3.5 TRAINING THE KNOWLEDGE ROUTER

The router $f_{\text{router}}$ is the only trainable component in our framework and is implemented with Qwen3-1.7B-Base. Its objective is to learn how to map $[D; C; Q]$ to minimal, neutral retrieval queries that correctly index candidate-specific knowledge, while leaving fine-grained symptom reasoning to the frozen LLM and the Answer stage.

We train the router via supervised distillation from a larger teacher model (Qwen3-235B-A22B). First, we construct a hierarchical encyclopedic instruction dataset based on the WIT corpus, containing structured descriptions $D$, candidate sets $C$, and teacher-generated retrieval instructions $q^*$ (details in Section 4.1). The teacher is prompted with $[D; C; Q]$ to produce high-quality, context-appropriate but feature-neutral queries such as "Retrieve symptom knowledge of [entity] on [organ]."

The student router is then optimized to imitate these instructions with a standard negative log-likelihood objective:

$$L(\theta) = -E_{(D, C, q^*) \sim \mathcal{D}}[\log P_\theta(q^* \mid [D; C; Q])],$$

where $\mathcal{D}$ is the distilled training set. To improve stability and generalization, we adopt a curriculum schedule that gradually shifts from coarse-grained to fine-grained entity distinctions, but the router is never required to produce detailed symptom descriptions itself. Instead, it learns a robust mapping from visual context and candidate identity to neutral retrieval templates.

All implementation details of dataset construction, stratified sampling, text-to-text Look simulation, and the two-stage distillation protocol are provided in Section 4.1.

## 4 EXPERIMENTS

### 4.1 DATA PREPARATION

To enable the Knowledge Router (Qwen3-1.7B-Base) to formulate discriminative search queries, we constructed a Hierarchical Encyclopedic Instruction Dataset containing 8,000 samples. We used the Wikipedia-based Image Text (WIT) dataset (Srinivasan et al., 2021) as the data source. WIT provides alignment between visual entities and encyclopedic metadata, which supports the learning of query generation patterns required for fine-grained classification.

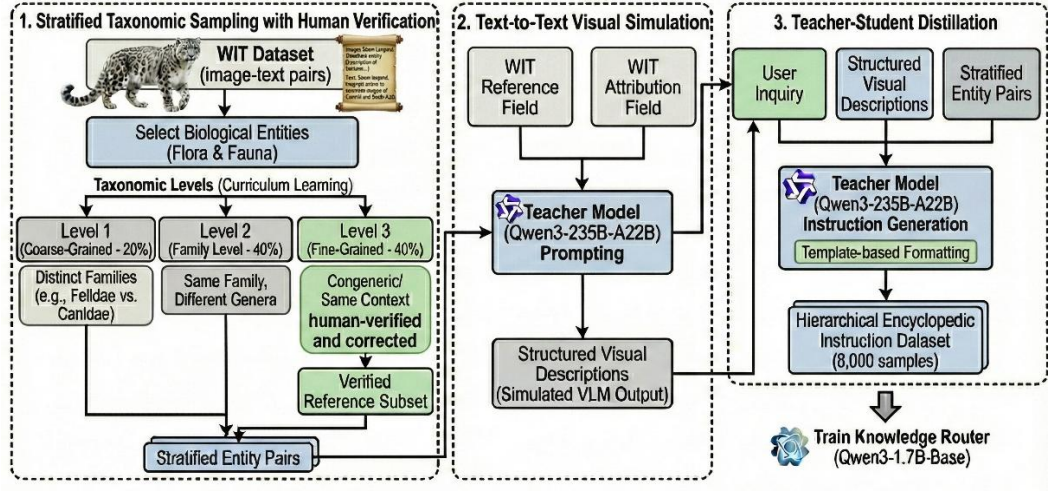

**1. Stratified Taxonomic Sampling with Human Verification**

We selected biological entities (Flora and Fauna) from WIT and organized them into three levels based on taxonomic distance. This structure implements a curriculum learning strategy:

- *Level 1 (Coarse-Grained Separation - 20%)*: Pairs are sampled from distinct taxonomic families (e.g., *Felidae* vs. *Canidae*) to establish baseline retrieval patterns.

- *Level 2 (Family-Level Distinction - 40%)*: Pairs share the same Family but differ in Genera, requiring intermediate discrimination.

- *Level 3 (Fine-Grained Disambiguation - 40%):* We selected congeneric species or entities appearing within the same Wikipedia context window. This level presents the highest task difficulty, requiring the model to identify specific visual attributes (e.g., *color distribution*) for disambiguation.

To ensure data quality, the Level 3 pairs were human-verified and corrected. Annotators reviewed a randomly sampled subset of 50 pairs to validate the accuracy of the generated instructions and correct any missing or misaligned attributes. They verified whether the extracted instructions correctly targeted the attributes necessary to distinguish the paired entities, yielding a human-curated reference subset for instruction tuning.

**2. Text-to-Text Visual Simulation**

To simulate the "Look" phase without visual encoding costs, we used Text-to-Text Simulation. (Hinton et al., 2015; Fang et al., 2021).We prompted the teacher model (Qwen3-235B-A22B) to combine the *Reference/ Alt and Attribution* fields from WIT into structured visual descriptions. This process creates a standardized input format that matches the output structure of the frozen VLM encoder.

**3. Teacher-Student Distillation**

We trained the router using instruction pairs generated by Qwen3-235B-A22B. Given a User Inquiry and Structured Visual Description, the teacher model generates a retrieval instruction specifying the knowledge required to distinguish between candidates (e.g., "Retrieve morphological profiles for [Species A] and [Species B], focusing on [Attribute X].").

To constrain the output space, we use template-based formatting. This simplifies the learning task for the 1.7B router, allowing it to focus on generating relevant retrieval keywords rather than arbitrary text variations.

## 4.2 EVALUATION DATASETS

We used AgroBench as the evaluation suite. AgroBench covers 203 crop categories and 682 disease categories, providing a comprehensive basis for assessing vision-language models in agricultural contexts. To balance evaluation scope with computational cost, we adopted a Two-Tier Evaluation

Strategy, grouping the seven benchmark tasks based on their representativeness and application scope:

**1. Core Task (Full Evaluation)**

We selected Disease Identification (DID) as the primary benchmark. As the most representative and widely applicable task in precision agriculture, DID involves distinguishing subtle morphological variations among disease categories. We evaluated the model on the complete DID test set (1,502 samples) to ensure a reliable assessment of its classification performance .

**2. Complementary Tasks (Random Sampling)**

For the remaining six tasks, we applied random sampling (50 samples per task). This approach enables efficient validation of model capabilities across diverse agronomic scenarios. These tasks include:

- Fine-Grained Recognition: Weed Identification (WID) and Pest Identification (PID).

- Agronomic Decision Making: Crop Management (CMN) and Disease Management (DMN).

- Operational Knowledge: Machine Usage QA (MQA) and Traditional Methods (TM).

This design enables rigorous evaluation of core competency (DID) while efficiently assessing generalization across the full spectrum of agricultural tasks.

4.3  PRE-TRAINED LANGUAGE MODELS S

We evaluated the feasibility of LOOK, RECITE, THEN ANSWER on the Qwen2-VL-72B model, but both the recite and answer stages employed Moonshot-Kimi-K2-Instruct (a model without visual capabilities). Additionally, the paper selected evaluation data from GPT-4o, Gemini 1.5-Pro, and human experts on AgroBench for comparison.

Due to space constraints, detailed introductions of these models are provided in Appendix .

4.4  EXPERIMENTS RESUALTS

Our framework achieves substantial improvements over baseline models across all evaluated tasks. Most notably, on Weed Identification (WID), we observe a +23.52 percentage point improvement over the `Qwen2-VL-72B` baseline (34.48% → 58.00%), representing a 68.2% relative gain—the largest absolute and relative improvement across all tasks. On Disease Identification (DID), our method achieves 78.03% accuracy, surpassing the baseline by +20.04 points (+34.5% relative). On Machine Usage QA (MQA), our framework achieves 94.00% accuracy—a +13.14 point improvement (+16.2% relative) over the baseline, representing the highest final accuracy among all tasks.

Compared to GPT-4o, which possesses a larger parameter scale than our baseline model `Qwen2-VL-72B`, our framework consistently outperforms it across all tasks: +13.83 points on WID (58.00% vs. 44.17%), +13.85 points on DID (78.03% vs. 64.18%), and +11.16 points on MQA (94.00% vs. 82.84%).

Task-Specific Analysis. The magnitude of improvement varies systematically with task characteristics:Fine-grained visual discrimination (WID, DID): The largest absolute and relative gains appear on tasks requiring fine-grained botanical and pathological discrimination, suggesting that candidate-guided knowledge retrieval is most effective when visual features alone are ambiguous and domain knowledge provides critical discriminative information.

Knowledge-intensive reasoning (MQA): Despite the smallest absolute improvement (+13.14 points), the highest final accuracy (94.00%) validates that explicit knowledge recitation outperforms implicit reasoning in retrieving procedural facts, and that the baseline model already demonstrates strong performance on this task.

This pattern aligns with our hypothesis: structured decoupling benefits tasks where (i) perception must remain objective to avoid confirmation bias, and (ii) specialized knowledge must be precisely activated rather than implicitly accessed..

|  | WID | DID | MQA |
|---|---|---|---|
| 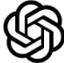 GPT 4o | 44.17 | 64.18 | 82.84 |
| **Our framework** | **58.00** | **78.03** | **94.00** |
| 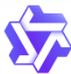 QwenVLM-72B | 34.48 | 57.99 | 80.86 |
| 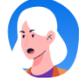 human | 20.00 | 25.00 | 57.50 |

### 4.5 ERROR ANALYSIS: WHERE DOES THE FRAMEWORK STILL FAIL?

Manual analysis of 50 DID errors reveals four failure modes:

| Error Type | Frequency | Root Cause |
|---|---|---|
| Ambiguous visual evidence | 38% | Multiple diseases exhibit similar symptoms in D |
| Candidate incompleteness | 30% | VLM fails to nominate correct answer in C |
| Temporal reasoning failure | 18% | Disease in early stage; characteristic symptoms not yet manifested |
| Context integration failure | 14% | Non-visual constraints in Q (e.g., planting date) ignored |

Key Insight: 68% of errors (Types 1-2) stem from limitations in the Look/Recite stages rather than the Answer stage. This suggests future work should prioritize improving candidate generation and visual description granularity over refining verification mechanisms.

Limitations on Ablation Studies. Due to computational resource constraints and project timeline limitations, we did not conduct systematic ablation experiments to isolate the marginal contribution of each module (Look, Recite, Answer). While the consistent cross-task improvements (+13-24 points) provide evidence for the effectiveness of our integrated design, we cannot definitively quantify whether performance gains arise additively from independent mechanisms or through synergistic interactions between stages. Additionally, our current evaluation is limited to Qwen2-VL-72B as the frozen VLM backbone; we have not validated the framework's generalizability across diverse VLM architectures (e.g., LLaVA, InternVL, Qwen3-VL). Future work with expanded computational resources should prioritize (i) controlled ablations to empirically validate our architectural hypotheses, and (ii) cross-architecture experiments to assess whether our decoupling strategy transfers to other foundation models with different visual encoders and training paradigms.

## 5 DISCUSSION

### 5.1 From Neural Gating to Semantic-Level Gating: Macro-Level Validation of Micro-Level Principles

Existing gating mechanisms operate at the element-wise or head-wise level, implementing input-dependent sparsity through scalar modulation. Qiu et al. (2025) systematically demonstrated that applying sigmoid gating after SDPA outputs achieves three key effects simultaneously: (1) breaking the low-rank linear bottleneck between $W_V$ and $W_O$ by introducing non-linearity; (2) enabling input-dependent sparsity through query-dependent sparse gate values; (3) fundamentally eliminating the attention sink phenomenon, as gate values can approach zero to effectively suppress outputs from irrelevant heads. These methods exhibit significant improvements in training stability and perplexity at scale, representing efficient and empirically validated architectural innovations.

Our Router reproduces similar gating logic at the conceptual level. The distinction is that the Router's decisions occur in an interpretable semantic space: it explicitly selects concrete concepts such as "foliar rust" or "fruit rust," rather than suppressing activations across thousands of incomprehensible hidden dimensions. This design sacrifices computational efficiency—the Router requires 1.7B parameters and explicit retrieval steps—but offers two advantages:

First, decision traceability. When the model fails at diagnosis, we can observe which incorrect disease candidates the Router selected or which irrelevant symptoms it retrieved, whereas the failure modes of neural gating remain hidden within thousands of activation dimensions.

Second, macro-level validation. The non-linearity and sparsity that Qiu et al. (2025) achieve through head-wise or element-wise gating within Transformers, we explicitly reproduce at the semantic level across the VLM reasoning chain. Experiments demonstrate that this explicit semantic routing effectively reduces reasoning-driven hallucinations on AgroBench, confirming that the principles of input-dependent sparsity and selective gating remain critical in cross-modal scenarios—even when implemented through a less efficient but more transparent approach.

In other words, our work does not aim to replace efficient neural gating, but rather uses a computationally expensive yet semantically interpretable version to demonstrate the necessity of such mechanisms in VLMs.

5.2 DeepSeek Sparse Attention (DSA) and the Lightning Indexer

When processing long contexts, DeepSeek V3.2 introduced DeepSeek Sparse Attention (DSA). The core component of DSA is the "Lightning Indexer"—a lightweight neural network that rapidly predicts the top-K most relevant Key-Value blocks before expensive attention computation.
The workflow proceeds as follows:

*Index*: The indexer computes relevance scores between queries and compressed key representations

*Select:* Top-K relevant blocks are selected

*Attend:* Fine-grained attention computation is performed only over the selected blocks

This "retrieve-then-attend" mechanism reduces computational complexity from quadratic to linear while improving the model's robustness to irrelevant noise by filtering out distractors.

Functional Analogy with Our Router: Our Router plays a role analogous to a "semantic-level Lightning Indexer" within the VLM reasoning chain. While DeepSeek's indexer filters relevant Key-Value pairs at the token level, our Router filters relevant disease candidates and symptom knowledge at the concept level. Both share a core architectural principle: use a lightweight module to rapidly locate the most relevant information subset before executing expensive reasoning computation. The distinction lies in the operational space: DeepSeek's indexer operates in implicit embedding space, whereas our Router operates in explicit semantic space, rendering the filtering logic interpretable and verifiable. This design choice again reflects our tradeoff between efficiency and interpretability: we sacrifice end-to-end computational efficiency in exchange for reasoning transparency.

REPRODUCIBILITY STATEMENT

**Evaluation datasets** The datasets used in our experiments are all publicly accessible.

**Prompts** We provide some used prompts in the appendix.

**Source code** Though the prompt examples in the appendix should be enough to reproduce all the results in our paper, we open-source all the evaluation code at
https://github.com/cxk66666666666/look-recite-answer-vlm.

REFERENCES

Zhiqiu Lin, Xinyue Chen, Deepak Pathak, Pengchuan Zhang, and Deva Ramanan. Revisiting the role of language priors in vision-language models. In International Conference on Learning Representations (ICLR), 2024

Haolang Lu, Boren Chu, Weiye Fu, Guoshun Nan, Junning Liu, Minghui Pan, Qiankun Li, Yi Yu, Hua Wang, and Kun Wang. Mitigating hallucination in multimodal reasoning via functional attention control. arXiv preprint arXiv:2510.10285, 2025.

Yunfan Gao, Yun Xiong, Xinyu Gao, Kangxiang Jia, Jinliu Pan, Yuxi Bi, Yi Dai, Jiawei Sun, and Haofen Wang. Retrieval-augmented generation for large language models: A survey. arXiv preprint arXiv:2312.10997, 2023.

Zhihan Zhang, Xin Luna Dong, and Heng Ji. Towards understanding systems trade-offs in retrieval-augmented generation. arXiv preprint arXiv:2412.11854, 2024.

Victor Weixin Liang, Yuhui Zhang, Yongchan Kwon, Serena Yeung, and James Y. Zou. Mind the gap: Understanding the modality gap in multi-modal contrastive representation learning. In Advances in Neural Information Processing Systems (NeurIPS), 2022.

Xinyue Chen et al. Learning to point visual tokens for multimodal grounded reasoning. arXiv preprint arXiv:2505.18842, 2025.

Xuweiyi Chen, Ziqiao Ma, Xuejun Zhang, Sihan Xu, Shengyi Qian, Jianing Yang, David F. Fouhey, and Joyce Chai. Multi-object hallucination in vision language models. In Advances in Neural Information Processing Systems (NeurIPS), 2024.

Jie Zhou, Mengdong Zhang, Nan Han, Hui-Feng Li, and Xin Yu. Multi-modal proactive reasoning with decoupled eyesight and wisdom. In Proceedings of the 2025 Conference on Empirical Methods in Natural Language Processing (EMNLP), 2025.

Jiuhai Chen et al. Self-rewarding vision-language model via reasoning-driven verification. arXiv preprint arXiv:2508.19652, 2025.

Jean-Baptiste Alayrac, Jeff Donahue, Pauline Luc, Antoine Miech, Iain Barr, Yana Hasson, Karel Lenc, Arthur Mensch, Katherine Millican, Malcolm Reynolds, et al. Flamingo: a visual language model for few-shot learning. In Advances in Neural Information Processing Systems (NeurIPS), 2022.

Meng-Jiun Chiou. Learning structured representations of visual scenes. PhD thesis, 2022.

Qiuhong Anna Shi et al. Activation control for efficiently eliciting long chain-of-thought in large language models. arXiv preprint arXiv:2505.17697, 2025.

Ting Zhang, Kun Yi, and Liang Zheng. Time-, memory- and parameter-efficient visual adaptation. In European Conference on Computer Vision (ECCV), 2024.

Che Jiang, Dequan Wang, Qiang He, Xinhao Li, Linkai Peng, and Yu Qiao. Freeze the backbones: A parameter-efficient contrastive approach to robust medical vision-language pre-training.

Meghana Luo et al. Visual description grounding reduces hallucinations and boosts reasoning in large vision-language models. arXiv preprint arXiv:2405.15683, 2024.

Qiuhong Anna Shi et al. A cognitive paradigm approach to probe the perception-reasoning decoupling in vision-language models. arXiv preprint arXiv:2501.13620, 2025.

Yiyang Li, Daniel Fried, and Eliana Lorch. Open-ended visual question answering guided by world knowledge. In Findings of the Association for Computational Linguistics: ACL 2023, pages 2321–2336, 2023.

Fabio Petroni, Patrick Lewis, Aleksandra Piktus, Tim Rocktäschel, Yuxiang Wu, Alexander H. Miller, and Sebastian Riedel. Language models as knowledge bases: On entity representations, storage capacity, and paraphrased queries. In Proceedings of the 16th Conference of the European Chapter of the Association for Computational Linguistics (EACL), 2021.

Alex Mallen, Akari Asai, Victor Zhong, Rajarshi Das, Daniel Khashabi, and Hannaneh Hajishirzi. When not to trust language models: Investigating effectiveness of parametric and non-parametric memories. In Proceedings of ACL, 2023.

Krishna Srinivasan, Karthik Raman, Jiecao Chen, Michael Bendersky, and Marc Najork. WIT: Wikipedia-based image text dataset for multimodal multilingual machine learning. In Proceedings of the 44th International ACM SIGIR Conference on Research and Development in Information Retrieval (SIGIR '21), pages 2443–2449, 2021.

Zhiyuan Fang, Jianfeng Wang, Xiaowei Hu, Lijuan Wang, Yezhou Yang, and Zicheng Liu. Compressing visual-linguistic model via knowledge distillation. In Proceedings of the IEEE/CVF International Conference on Computer Vision (ICCV), 2021.

Zihan Qiu, Zekun Wang, Bo Zheng, Zeyu Huang, Kaiyue Wen, Songlin Yang, Rui Men, Le Yu, Fei Huang, Suozhi Huang, Dayiheng Liu, Jingren Zhou, and Junyang Lin. Gated attention for large language models: Non-linearity, sparsity, and attention-sink-free. In Advances in Neural Information Processing Systems (NeurIPS), 2025.

DeepSeek-AI. Introducing DeepSeek-V3.2-Exp. DeepSeek Official News, September 2025. Available at https://api-docs.deepseek.com/news/news250929.

Alibaba Cloud AI Research. ZeroSearch: Zero-cost search engine simulation for reinforcement learning. Technical Report, 2025. Available at https://github.com/alibaba/ZeroSearch and Hugging Face.

Zhiqing Sun, Xuezhi Wang, Yi Tay, Yiming Yang, and Denny Zhou. Recitation-augmented language models. In International Conference on Learning Representations (ICLR), 2023.

## A  MODEL OUTPUTS

根据您提供的图像描述和权威检索信息，结合植物学特征分析，覆盖池塘水面的植物是菱角（菱）与浮萍的混合群落，具体识别如下：

## 1. 主体植物：菱角（学名：*Trapa* spp.）

- 核心特征：
  - 深绿色、近圆形或菱形叶片（叶片边缘常呈锯齿状，但整体轮廓近圆形），叶片漂浮于水面；
  - 叶柄中部膨大形成气囊结构，帮助植物漂浮；
  - 果实为可食用的菱角（成熟后呈四角形，俗称"水栗"），但未成熟时叶片密集覆盖水面。

- 为何是菱角？
  - 图像中"深绿色、圆形叶片""密集覆盖水面""需人工清理"等描述高度匹配菱角特性；
  - 菱角是中国南方池塘、湖泊的典型富营养化水体植物，生长迅速时会堵塞水面，需定期打捞（如图中男子用船清理的行为）；
  - 检索验证显示，中国农村水体中"圆形浮叶植物"最常见且符合描述的即为菱角（如 *Trapa bispinosa* 或 *Trapa natans*）。

## 2. 夹杂的亮绿色小片植物：浮萍（学名：*Lemna minor* 等）

- 核心特征：
  - 小型漂浮植物，单个个体仅1-5毫米，呈椭圆形或椭圆形；
  - 亮绿色、无根或具细根，常成片密集漂浮，形成"斑驳绿毯"；
  - 繁殖极快，易在富营养化水体中爆发，但本身无经济价值，多被视作水质污染指示植物。

- 为何是浮萍而非藻类？
  - 检索结果明确指出：图像中"亮绿色小片"实际为浮萍（被子植物），而非藻类（低等植物，如绿藻通常呈丝状或胶状）；
  - 浮萍与菱角共生性极强，在静水池塘中常共同生长，且浮萍的亮绿色与菱角的深绿色形成明显对比（符合"斑驳绿毯"描述）。

## 关键区分点：避免常见误解

| 植物类型 | 菱角 | 浮萍 | 水葫芦 | 睡莲 |
|---|---|---|---|---|
| 叶片形状 | 近圆形/菱形，边缘略锯齿 | 卵圆形，单片极小（<5mm） | 椭圆形，叶柄膨大 | 圆形，有V形缺口 |
| 根系 | 叶柄有气囊，根系退化 | 无根或细根 | 长须根 | 水下根系发达 |
| 是否可食用 | 果实（菱角）可食用 | 不可食用 | 果实可食用（但需处理） | 花可观赏，根茎可食 |
| 图像匹配度 | ✅深绿、密集、需清理 | ✅亮绿、细小、成片 | ❌叶片较大且叶柄膨大 | ❌叶片有缺口，花显著 |

🌱 结论：
图像中覆盖池塘的植物是菱角与浮萍的混合群落。

- 大片深绿色圆形叶片 → 菱角（需人工打捞的经济作物，但过度繁殖会破坏水体生态）；
- 细小亮绿色漂浮物 → 浮萍（常见于富营养化水体，生长迅速但无经济价值）。

这种组合在中国南方乡村池塘中极为典型，反映了人与自然互动的日常场景（如男子清理水体以维持通航和生态平衡）。若需进一步确认，可观察是否有菱角果实（成熟时呈四角形）或浮萍的细小根系（显微镜下可见）。

分享自 agrovista

清言 你的全能伙伴

用户_m1I6G8 的提问

## 这是什么植物

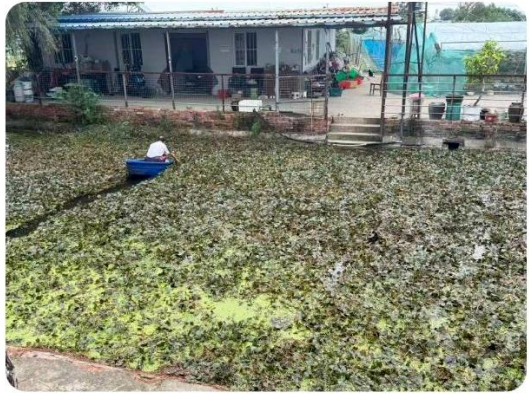

从图片来看，水面大量覆盖的植物很可能是**浮萍**（属于浮萍科浮萍属的漂浮植物）。浮萍常在水体富营养化或静水环境下大量繁殖，形成密集的漂浮层，符合图中植物的特征。

当然，若要更精准鉴定，需结合植物细节（如叶片形状、是否有根等），但从常见水生植物群落和场景判断，"浮萍"是最可能的答案~

AI生成

Qwen3-VL-235B-A22B

这是什么植物

Qwen3-VL-235B-A22B

已完成思考

81,920 tokens 预算

图中覆盖水面的植物是**菱角**（学名：Trapa bispinosa，菱科菱属水生草本植物）。

其典型特征为：叶片呈盾状、浮于水面，密集覆盖水域，植株形态与池塘种植环境高度吻合。菱角是重要的水生经济作物，果实"菱角"可食用，图中水面作业场景也符合菱角采摘的常见方式。

若需进一步区分，可注意其叶片边缘有细锯齿、叶柄中部膨大（图中因密集覆盖不易直接观察），与浮萍、水葫芦等植物有明显差异。

图像编辑　 WebDev　 图像生成　 视频

有什么我能帮您的吗？

　 深度思考 　 搜索　

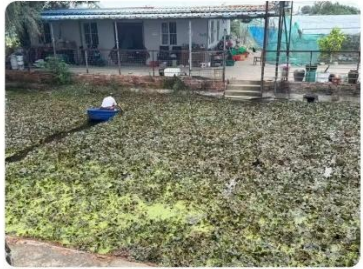

这个是什么植物

从照片来看，这片水面上布满的植物是一种典型的水生浮叶植物，看起来像是 **菱角（Trapa natans，俗称菱）**。

特征可以对上：

- 叶片呈菱形或三角形，边缘有小锯齿；
- 叶子集中漂浮在水面，像是一大片绿色的"浮毯"；
- 常见于池塘、湖泊、人工水域；
- 菱的叶片往往秋季开始发黄、枯萎，就像图里这种颜色。

前景下方的淡绿色小点状植物，则是 **浮萍（Lemna 属）**，它常和菱、睡莲等一起出现。

总结：

- 主体大面积的植物是 **菱角（Trapa natans）**。
- 淡绿色覆盖物是 **浮萍**。

要不要我帮你写一个小科普，介绍菱在农业/水产中的用途和生态影响？

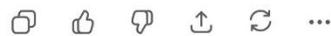

询问任何问题

ChatGPT 也可能会犯错。请核查重要信息。

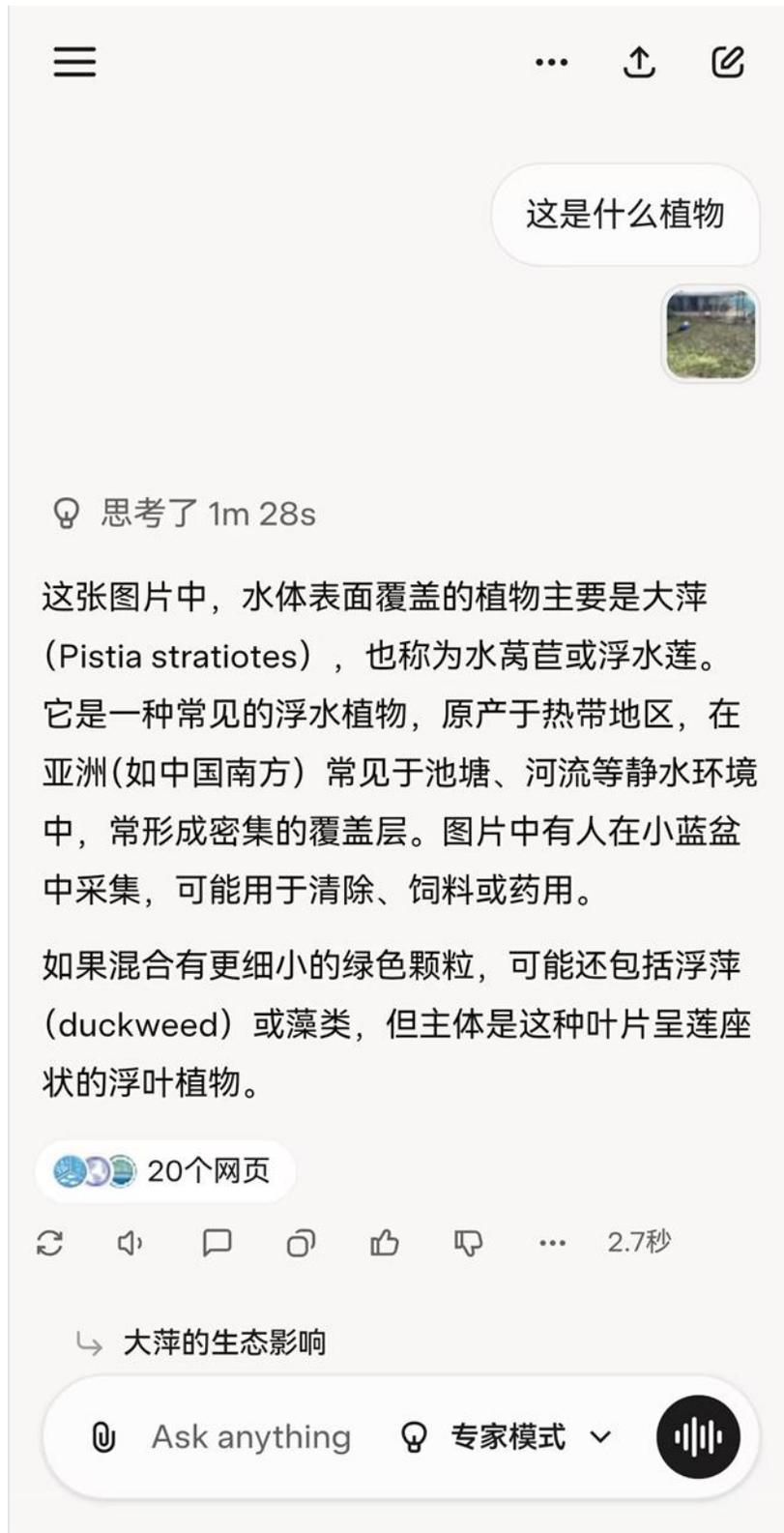

Fig. Images comparing our framework with GLM-4.5, Qwen3-VL-235B-A22B, GPT-5, and Grok Expert.

## B  PROMPT DESIGNS

> *You are a professional plant pathologist. Please provide a detailed and objective description based entirely on the image content. Follow these principles:*
>
> *Distinguish subject from background: Focus your analysis on the plant subject in the image, ignoring irrelevant background interference.*
>
> *Combine global and local perspectives: First describe the overall condition of the plant (global features), then carefully examine and describe any suspicious local lesion features in detail. When describing lesions, you must pay special attention to their morphological characteristics, using precise geometric or morphological terminology, such as whether the lesions exhibit concentric ring patterns, wavy edges, striped patterns, irregular shapes, etc. Do not make any guesses or associations.*

> *You are a "hypothesis generator" for plant pathology diagnosis. Your task is to generate a list of diagnostic hypotheses containing multiple possibilities based on preliminary, objective image descriptions and the original question, and provide clear guidance for the next step of visual verification. Please strictly follow this workflow:*
>
> *Entity extraction and retrieval: Extract the plant name and core symptoms, construct "plant + symptom" combinations for retrieval.*
>
> *Generate hypotheses and verification checklist: Output a hypothesis checklist.*

> *You are a top-tier plant pathology diagnostic expert with extensive field experience, skilled at conducting prudent logical reasoning under conditions of incomplete information. Your task is to integrate all preliminary investigation information to form a professional, rigorous, and responsible diagnostic report.*
>
> *Please strictly follow this advanced analytical framework:*
>
> *Information synthesis: Integrate the original question, objective description, hypothesis checklist, and visual verification report.*
>
> *Advanced diagnostic reasoning:*
>
> *Evidence weight assessment: Dynamically evaluate hypothesis probabilities based on verification results.*
>
> *Holistic view and distribution pattern analysis: Determine whether occurrence is sporadic (biotic stress) or widespread (physiological stress).*
>
> *Causal chain reasoning: Consider "invisible pathogens" (such as root diseases causing leaf wilting).*
>
> *Structured diagnostic conclusion and action plan:*
>
> *Primary diagnosis and confidence level.*
>
> *Secondary possibilities.*
>
> *Diagnostic limitations: Clearly identify information gaps.*
>
> *Questions and verification steps: Propose follow-up examination recommendations.*
>
> *Preliminary management recommendations.*

Table. the prompts for the Look, Router, and Answer stages

## C  PRE-TRAINED LANGUAGE MODELS

**QwenVLM-72B**

`Qwen2-VL-72B`（QwenVLM-72B）is a large-scale vision-language model developed by Alibaba Cloud, part of the Qwen (Tongyi Qianwen) series. With 72 billion parameters, it is designed to process both visual and textual inputs, enabling multimodal understanding tasks such as image captioning, visual question answering, and visual reasoning. The model employs a transformer-based architecture that integrates visual encoders with language modeling capabilities, allowing it to generate contextually relevant responses based on image content.

**Moonshot-Kimi-K2-Instruct**
Moonshot-Kimi-K2-Instruct is a text-only large language model developed by Moonshot AI (Dark Side of the Moon Technology). This instruction-tuned model is optimized for following complex instructions and generating coherent, contextually appropriate responses. Notably, it does not possess visual processing capabilities, making it suitable for the recite and answer stages of our framework where visual information has already been extracted and converted to text. The model demonstrates strong performance in reasoning, knowledge retrieval, and natural language generation tasks.

**GPT-4o**
GPT-4o (where "o" stands for "omni") is OpenAI's multimodal large language model capable of processing text, images, and audio inputs. Released as an advancement over GPT-4, GPT-4o features enhanced multimodal integration, improved reasoning capabilities, and faster response times. In the context of AgroBench evaluation, GPT-4o serves as a strong baseline for vision-language understanding tasks, particularly in specialized domains requiring both visual perception and domain knowledge.

**Gemini 1.5-Pro**
Gemini 1.5-Pro is Google DeepMind's advanced multimodal AI model featuring an extended context window and enhanced reasoning capabilities. It processes text, images, video, and audio inputs within a unified architecture. The model incorporates a mixture-of-experts (MoE) architecture for efficient scaling and demonstrates strong performance across diverse tasks including visual understanding, complex reasoning, and long-context processing. Its inclusion in AgroBench evaluation provides a comparative benchmark from a leading multimodal foundation model